\documentclass[journal]{IEEEtran}
\usepackage{amsmath,amsfonts}
\usepackage{algorithmic}
\usepackage{algorithm}
\usepackage{array}
\usepackage[caption=false,font=normalsize,labelfont=sf,textfont=sf]{subfig}
\usepackage{textcomp}
\usepackage{stfloats}
\usepackage{url}
\usepackage{verbatim}
\usepackage{graphicx}
\usepackage{xcolor} 
\usepackage{soul}
\usepackage[backend=biber,style=numeric-comp,sorting=none]{biblatex}
\bibliography{bibtex/main}
\begin{document}

\title{Accelerating evolutionary exploration through language model-based transfer learning}

\author{Maximilian Reissmann$^*$, Yuan Fang, Andrew S. H. Ooi and Richard D. Sandberg
\\
Department of Mechanical Engineering, The University of Melbourne
\thanks{Manuscript created October 2023; The authors are with the Department of Mechanical Engineering, The University of Melbourne Grattan Street, Parkville Victoria, 3010, Australia}
\thanks{*Maximilian Reissmann: e-mail: reissmannm@student.unimelb.edu.au}}

\markboth{Journal of \LaTeX\ Class Files,~Vol.~14, No.~8, August~2021}%
{Shell \MakeLowercase{\textit{et al.}}: A Sample Article Using IEEEtran.cls for IEEE Journals}

\IEEEpubid{0000--0000/00\$00.00~\copyright~2021 IEEE}
\maketitle

\begin{abstract}
Gene expression programming is an evolutionary optimization algorithm with the potential to generate interpretable and easily implementable equations for regression problems. Despite knowledge gained from previous optimizations being potentially available, the initial candidate solutions are typically generated randomly at the beginning and often only include features or terms based on preliminary user assumptions. This random initial guess, which lacks constraints on the search space, typically results in higher computational costs in the search for an optimal solution. Meanwhile, transfer learning, a technique to reuse parts of trained models, has been successfully applied to neural networks. However, no generalized strategy for its use exists for symbolic regression in the context of evolutionary algorithms. In this work, we propose an approach for integrating transfer learning with gene expression programming applied to symbolic regression. The constructed framework integrates Natural Language Processing techniques to discern correlations and recurring patterns from equations explored during previous optimizations. This integration facilitates the transfer of acquired knowledge from similar tasks to new ones. Through empirical evaluation of the extended framework across a range of univariate problems from an open database and from the field of computational fluid dynamics, our results affirm that initial solutions derived via a transfer learning mechanism enhance the algorithm's convergence rate towards improved solutions.
\end{abstract}

\begin{IEEEkeywords}
Gene Expression Programming, Transfer Learning, Symbolic Regression
\end{IEEEkeywords}

\section{Introduction}
\IEEEPARstart{E}{volutionary} optimization techniques such as Hollands genetic algorithm \cite{558db5e0-7a5a-3da1-8b88-2f13b93dbaf1}, Koza's genetic programming \cite{koza1992genetic} or the Gene Expression Programming introduced by Fereirra \cite{ferreira2006gene} offer powerful strategies for symbolic regression that enable the extraction of robust and interpretable models from data. 
These methods constitute population- and non-gradient-based metaheuristics where the optimization step of the candidate solutions emulates the evolution of species described by Darwin. They have been successfully applied in different engineering applications, like material science \cite{10.1002/jcc.27043}, \cite{wang_wagner_rondinelli_2019}, turbulence modeling \cite{Weatheritt2016}, \cite{Akolekar2021} or in domains like finance \cite{10.1024/1421-0185/a000241}, \cite{ 10.3390/app12136661} and healthcare \cite{10.1152/advan.00233.2020}, \cite{10.22159/ajpcr.2022.v15i8.45019}.
However, these strategies are often criticized for their non-deterministic behavior, which leads to unpredictable convergence times. This issue becomes particularly critical in environments where fitness evaluations are cost-intensive. Furthermore, the time required for convergence increases with the number of features in the model, limiting the practicality of these methods in solving complex, large-scale data problems. 
Driven by the motivation of reducing the exploration time inherent in evolutionary algorithms for symbolic regression, various methodologies have been proposed. These include a variety of genetic operator adaptations \cite{Zhong2017GeneSurvey}, feature selection techniques via an inductive bias aiming for a reduction of the dimensionality of the search space \cite{Cranmer2020DiscoveringBiases}, as well as strategies employing gradients or regularization to guide the search process more effectively \cite{ Petersen2019DeepGradients}, \cite{Brunton2016DiscoveringSystems}. Apart from only revising the symbolic regression itself, hybrid methodologies have emerged that integrate evolutionary search mechanisms with neural network feedback, facilitating the refinement of individual candidate solutions or the recreation of entire populations for a more efficient search \cite{Mundhenk2021,Zhang2021,li2023turbulence}. 

A different type of approaches are those that can be categorized as pre-trained or inductive. They use a retention mechanism (large language model) trained on a wide range of diverse mathematical problems, allowing a short inference in terms of suggesting an approximation for a particular problem sought \cite{Biggio2020ARegression, Kamienny2022End-to-endTransformers, Valipour2021, Vastl2022SymFormer:Architecture}. Notwithstanding the advantages of pre-trained language models in symbolic regression, these are characterized by substantial computational requirements. Specifically, they exhibit a high amount of parameters, necessitating an extensive training data set ($\geq 10^5$), and require significant hardware resources during the training phase.\IEEEpubidadjcol

A further technique that seeks to reduce the exploration time is referred to as transfer learning, whereby the application of the acquired knowledge from prior similar approximations takes into account the relationship between different domains. Transfer learning (TL) refers to a methodology transmitting knowledge from a source to a target task or across different domains \cite{Pan2010ALearning}, \cite{10.1016/j.neunet.2019.01.012}. The procedure is inspired by the role model from nature, whereby solving a novel complex task often leverages previously acquired knowledge from analogous challenges. This alignment with biological learning mechanisms highlights that problem-solving is rarely an isolated event but rather an iterative process that benefits from prior experiences.

Approaches such as \cite{Dinh2015TransferProgramming}, \cite{Haslam2016}, \cite{Muller2019}, or \cite{Munoz2020TransferProgramming} indicate a beneficial impact by transferring knowledge from former optimizations to more complex problems. Specifically, Dinh et al. \cite{Dinh2015TransferProgramming} proposed methodologies, including FullTree, SubTree, and BestTree, which involve transferring varying proportions or specific segments of expression trees from the final generation as a form of knowledge. Building on these, subsequent research by Haslam et al. \cite{Haslam2016} and Muller et al. \cite{Muller2019} sought to refine the measurement of how much knowledge to transfer to new problems while maintaining sufficient exploratory capabilities. Expanding beyond single-problem focus, O'Neil et al. \cite{ONeill2017CommonProgramming} investigated the abilities of the most common subtrees in the final generations of two previous optimizations. Munoz et al. \cite{Munoz2020TransferProgramming} applied constructive induction of features, expanding the feature space from the source to the target problem to solve the feature alignment problem.
To further optimize the selected proportion, Chen et al. \cite{Chen2019}, \cite{Chen2022} proposed a technique, weighting the instances for the new task and performing differential optimization to determine the most appropriate proportion by progressively testing them, whereby \cite{Chen2022} utilized an improvement regarding estimating a more favorable initialization and tendency to mitigate dominance from the introduced individuals.

More recently, Chu \cite{Chu2022TransferProgramming} proposed an approach that applies a semantic approximation technique using the best solutions of the source task to create new ones based on their semantic information, therefore building a kind of library and accounting for a common occurrence of subtrees. 

Outside the domain of symbolic regression, studies like \cite{8789920}, \cite{10.1007/978-3-030-64984-5_12}, or \cite{Wild2022} further have demonstrated the beneficial effect of using this methodology on combinatorial problems or program synthesis tasks. For instance, Ardeh et al. \cite{8789920} employed the knowledge from the most frequently occurring subtrees, which were extended to a probabilistic prototype tree in the subsequent research \cite{10.1007/978-3-030-64984-5_12}. Here, the probability of the nodes was fitted, accounting for parts of the best-performing individuals from a final generation. Moreover, the work of Wild et al. \cite{Wild2022} investigated the potential within the field of program synthesis using a deep neural network to identify potential candidates for a knowledge transfer.

Focusing on symbolic regression, the studies \cite{Dinh2015TransferProgramming}-\cite{Chen2022} aim to decrease the error and reduce the exploration time. There is always an immediate temporal and local connection between the approximation of the former optimization task and the task that needs to be solved. Furthermore, the knowledge acquired is expressed as trees, and the problem is that there is no representation to store it effectively. Finally, with the exception of \cite{Munoz2020TransferProgramming}, their test setup is often limited to uni- or bi-variate problems utilizing artificial data sets. 

Inspired by the idea of how language models preserve the acquired knowledge of learned equations and the global optimization capabilities of an evolutionary algorithm, this study aims to introduce a procedure to combine a language model with an evolutionary method. The composition enables transfer learning ability on different symbolic regression problems, focusing on the following aspects:
\begin{enumerate}
\item By using a language model, we aim to preserve the information from the most common subtrees.
\item An encoding mechanism facilitates a latent representation connecting the spatial information from the domain to the structure of the sequence.
\item The synergy between the language model and the applied evolutionary method, particularly the gene encoding, enables short and resource-efficient training. 
\end{enumerate}
To assess the capabilities of the proposed method, we employ an experimental design that explores the impact on multivariate regression problems from the UCI data set \cite{DuaGraff2017} and a duct flow scenario from the area of computational fluid dynamics. Calculating the flow of the duct requires an approximation of the mathematical relationship between the mean flow and turbulence-related quantities.
This paper is organized as follows. Section \ref{method} proposes the details of the adapted language model and the procedure for sampling a biased start population. A comparison of the standard GEP with the technique presented here demonstrates the influence on the exploration, underlining the iterations needed to approximate a solution. Section \ref{conclude} summarizes our findings and outlines future ideas.

\section{Methodology}
\begin{figure}[!t]
\centering
\includegraphics[width=3.2in, height=3.2in]{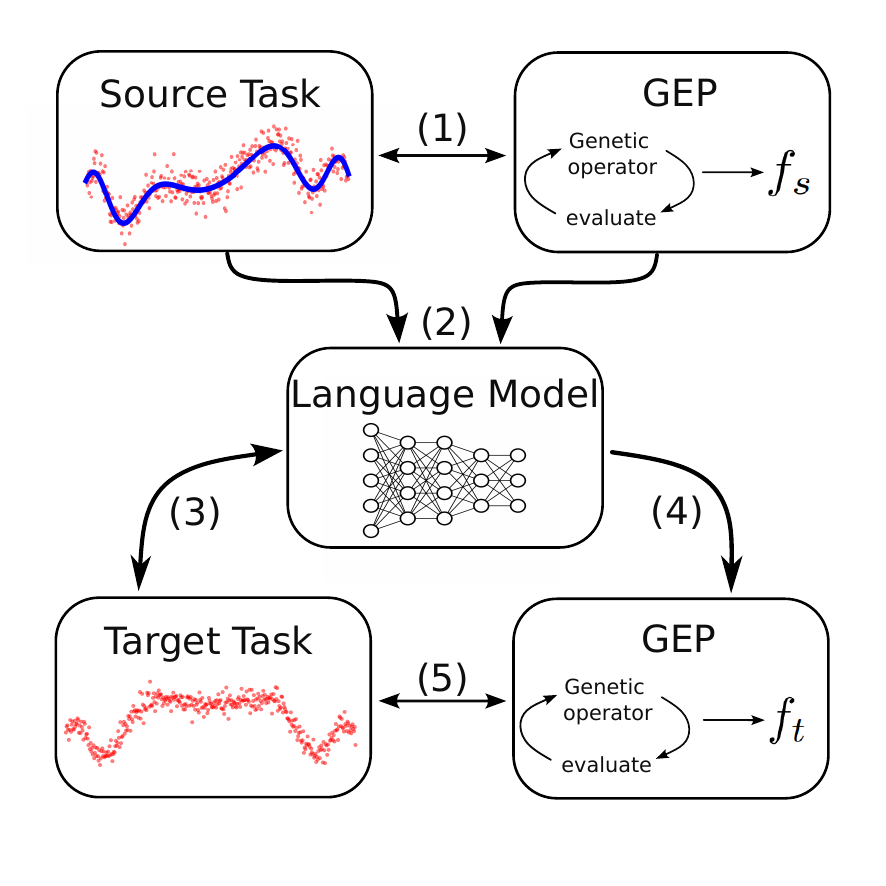}
\caption{Figure showing the interaction between the evolutionary method Gene Expression Programming (GEP) and a language in the context of transfer learning. The process is structured in five distinct steps. Initially, GEP is employed to approximate a function for a source task (1). This is followed by the training of the language model using the combination of a representation and the approximation (2). For the application to a subsequent problem (3)-(4), a proportion of the start population is generated using the case-related representation followed by an optimization (5).}
\label{fig_1:overview}
\end{figure}
\label{method}
\noindent This section describes the core components and their intercommunication to create a biased start population by using acquired knowledge from different source tasks. We first present an overview of how the language model interacts with the evolutionary method in combination with introducing relevant terminology. Followed by an outline of the idea behind the initialization by reflecting a relevant property of the GEP. Subsequently, the encoding and decoding procedure is explained by considering the interface to the evolutionary method. 

\subsection{Conceptual Overview}
\noindent Following the definition of \cite{Chen2022DiscoveringInvariance} or \cite{Dinh2015TransferProgramming}, the technique of transfer learning can be formally described in terms of the domain ($\mathcal{D}$) and task $(\mathcal{T})$ considerations. Here, a domain consists of a data space $\mathbf{X}$ and a probability distribution $P(\mathbf{X})$ governing all features and their values, mathematically expressed as $\mathcal{D}=\{\mathbf{X}, P(\mathbf{X})\}$. The term task, on the other hand, is essentially a function mapping from the given data space $\mathbf{X}$ to an output space $\mathbf{Y}$, defined as $\mathcal{T}=\{\mathbf{Y},f(.)\}$, where $f$ marks a decision function. For the application, it is assumed that knowledge acquired in the form of subtrees, specifically substrings within the gene from the explored function $f_s$, while addressing a source task $\mathcal{T}_s$ within a corresponding source domain $\mathcal{D}_s$ can beneficially inform the process approximating a target function $f_t$. 

A projection on how to use this concept on the GEP in combination with a language model is sketched in Fig. \ref{fig_1:overview} within five distinct stages. For a given problem (or a set of optimization problems), defined as finding $f_s$ with the objective $f_s(\mathbf{X})\approx\mathbf{Y}$ the evolutionary method is applied (1). Here, within the cause of the evolutionary exploration, each assessment of an individual yields a tiny amount of information on how a function for a problem can be potentially approximated. After encountering termination, a language model (2) is trained based on a set of the best approximations in combination with a representation of the $\mathcal{T}_s$. In detail, these expressions are tokenized based on their linear gene representation, preserving the structure of all their subtrees. For subsequent problems, the optimization method queries within the initialization process the language model based on the representation of $\mathcal{T}_t$ (3). This mechanism creates expressions in their linear representation and infers them into the GEP framework (4), which subsequently approximates a function $f_t$ (5).

\subsection{Initialisation}
\noindent Typically, the generation of candidate solutions in an evolutionary method, each represented as an equation (or individual), is governed by a rudimentary probabilistic model. Consider a finite set $C$ comprising $k$ symbols (here also referred to as tokens) and represented as $C=\{c_1, c_2,..., c_k\}$ (union set of terminal and nonterminal symbols $C_{nt}\cup C_{t}$). When performing symbol selection from the given set $n$ times, under the assumption that each sign has an equal likelihood of being chosen, the probability for any particular symbol $c_i$ follows a discrete and independent uniform distribution: $P(c_i)=\frac{1}{k}$. Constraints on this uniform distribution may arise from various factors, such as the depth of the calculated tree as discussed by Koza \cite{koza1992genetic}, or the symbol's position within the string as elaborated by Ferreira \cite{ferreira2006gene}. The aforementioned probabilistic formulation is notably insensitive to the context; it neither contains information about previously sampled symbols nor integrates any meta-information related to the specific optimization problem under consideration. The extension proposed below can be derived from the idea of autoregressive sequence generation from language modeling \cite{Mundhenk2021}, \cite{Petersen2019DeepGradients} under a given bias introduced by a problem representation \cite{, Kamienny2022End-to-endTransformers},\cite{Valipour2021}, \cite{Biggio2021NeuralScales} (through an encoder-decoder logic). Here, the selection is defined as:
\begin{equation}
\label{prop_eq}
P(c_t)=P(c_t|c_{t-1}, c_{t-2},...,c_{0};\theta)
\end{equation}
with $c_t$ denoting a particular token at position $t$ depending on a state $\theta$ provided by a model, which introduces information from the target task. 

\subsection{Gene Expression Programming}
\noindent Apart from the iterative adaptation, containing operators like selection, mutation and crossover, the GEP approach distinguishes itself from Koza's Genetic Programming \cite{koza1992genetic} (GP) and Holland's Genetic Algorithms \cite{558db5e0-7a5a-3da1-8b88-2f13b93dbaf1}. Specifically, Ferreira's approach \cite{ferreira2006gene} utilizes a linear scheme of fixed length to encode the symbolic expression, leading to a separable genotype and phenotype representation, referred to as Karva string. The sequence is partitioned into a head segment ($H$) and a tail segment ($T$), whereby the length is defined as ($2\cdot|H|+1$). The head is permitted to comprise both terminal and nonterminal symbols, while the tail is constrained to only consist of terminal symbols given by:
\begin{equation}
\label{eq-seq_prop}
\ln(P(c_t|c\in C_{NT} \land t \geq |H|)) \approx  -\infty \,.
\end{equation}
The Karva string is converted into a tree structure (phenotype), enabling recursive evaluation. This representation is able to create hierarchical structures similar to the canonical GP for approximating complex problems while consistently ensuring syntactic validity and mitigating the bloating effect likely in linear Genetic Programming (LGP) \cite{bloatGP}. For a deeper understanding of how the GEP and the particular operator are utilized, the reader is referred to Ferriera's work \cite{ferreira2006gene}.

\subsection{Tokenization}
\begin{figure}[!t]
\includegraphics[width=3.75in]{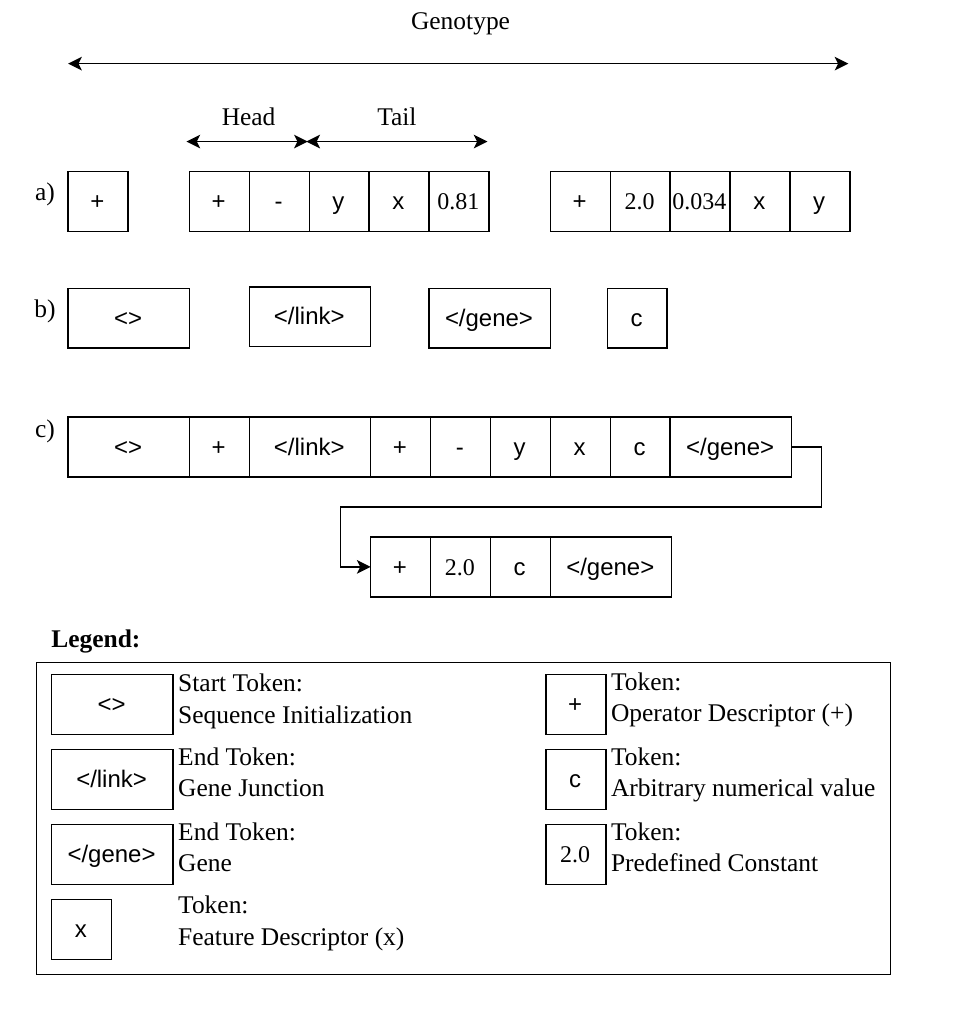}
\caption{Picture of the various elements of tokenization for an example of a genotype encoded by the gene expression programming method. Here, a) shows the original encoding of two genes with the labeling of the individual parts (Head, Tail). The rows b) and c) additionally depict the tokens introduced and the sequence formulated with these elements. The legend signifies the meaning of the individual tokens.}
\label{fig_tokenized_seq}
\end{figure}
\label{chap-token}
\noindent Within the domain of language modeling, a token describes the smallest fragment of a sequence for the input of the neural network \cite{Rai2021StudyTokenization}. For the given task of symbolic regression, the tokenization takes place on the character level. Hence, a token corresponds to a symbol from the set $\mathbf{C_{nt}} \cup \mathbf{C_{t}}$, where this is referred to as the vocabulary. Within the subsequent paragraphs, we use the terms symbol and token interchangeably. 

An example of the transformation of a genotype string into a tokenized sequence is visualized in Fig. \ref{fig_tokenized_seq}. Here, (a) shows two genes with a connecting token (`+'). In contrast, row (c) presents the resulting string under embedding a set of control tokens (b) and converting floating point numbers into a symbol (`c'), except for pre-defined constants like $2.0$. Notably, we only tokenize the part of the equation that is evaluated in their linear representation. Furthermore, the insertion of the control elements permits the sequence to be segmented according to further logical considerations. This, later on, allows steering of the prediction such that the sequence generation is less dependent on the number of genes or the length of a particular gene. Due to the various syntactic applications of the conjunction operators, on the one hand, linking distinct gene instances and, on the other hand, concatenating other tokens, they are also separated by a unique token ($</link>$). A second point deals with the discretization of the representation of numbers. There are solutions like that of Kamienny et al. \cite{Kamienny2022End-to-endTransformers}, allowing a comprehensive representation of floating point numbers but adding more tokens to the vocabulary and thus increasing the learning effort. Consequently, we decide on the variant of representing a skeleton \cite{Biggio2020ARegression}, \cite{Valipour2021} by introducing a small number of placeholder tokens for different numerical values inserted at inference time.

\subsection{Language model architecture}
\noindent The framework designed to enable a relational knowledge transfer incorporates an encoder-decoder architecture from the field of language modeling, especially the transformer introduced by Vaswani et al. \cite{Vaswani2017AttentionNeed}. The first part of the framework is responsible for capturing essential features and relationships of the input values and mapping them to a latent space to yield a fixed-size representation. The decoder, in contrast, is tasked with converting this latent information back into a coherent and interpretable sequence. This is accomplished by first computing a probabilistic vector and then applying a sampling technique to retrieve the subsequent symbol. Both are constructed from multiple transformer blocks that implement functionalities such as self-attention, layer normalization, and feed-forward neural networks. Several hyperparameters govern the size of the language model: the context window ($size_{w}$), the quantity and dimensionality of attention heads, the number of transformer blocks, and the dimension of the embedding ($d_{emb}$).
\\\\
\noindent\textbf{Encoder}: An encoder $(\mathcal{E})$ can be specified as a function that maps an input $\mathbf{X} = \{\mathbf{x}_1, \mathbf{x}_2,..., \mathbf{x}_i\},\,\mathbf{Y} = \{y_1, y_2,..., y_i\}$ into latent representation $\mathbf{\Tilde{z}}$:
\begin{equation}
\label{eqn-encoder}
\mathbf{\Tilde{z}} = \mathcal{E}(\mathbf{X},\mathbf{Y}; \theta_{\mathcal{E}}).
\end{equation}
Here, $\theta_{\mathcal{E}}$ denotes learnable parameters governing the encoder. The latent representation $(\mathbf{\Tilde{z}})$ encapsulates the essential features and relationships among the elements of $(\mathbf{X},\mathbf{Y})$ into a vector with a fixed size of $d_{emb}$, thus serving as a compressed version of the given task. Considering the input parameters for a generic symbolic regression algorithm and building on previous studies in which the transformer architecture has been employed \cite{ Kamienny2022End-to-endTransformers}, \cite{Valipour2021}, \cite{Biggio2021NeuralScales} we emphasize the requirement of the output of $\mathcal{E}$ being order-invariant to the input $(\mathbf{X},\mathbf{Y})$. In addressing this task, the literature provides different approaches like a set transformer \cite{lee2019set} (modified transformer block) or the pointNet \cite{Qi2017}. 

We employ the latter mentioned for equation (\ref{eqn-encoder}), particularly the transformation-net component (T-net) similar to the study \cite{Valipour2021} since there is no restriction on the input amount. The input to $\mathcal{E}$ is a matrix of dimension $n \times m$, where $m$ is the sum of the number of input features $\mathbf{X}$ and the corresponding output values $\mathbf{Y}$. The dimension $n$ represents the quantity of data points relevant to the problem under investigation. To facilitate the training of an encoder for multivariate tasks, it is requisite that $m$ is defined as greater than one; zeros can be conveniently padded during inference where certain features are absent. Algorithm (\ref{alg:tnet}) illustrates the mapping of the concatenation of $(\mathbf{X},\mathbf{Y})$ into $(\mathbf{\Tilde{z}})$, while also annotating the dimensions of the resultant vector, with the batch size ($b$), after each step. 
\begin{algorithm}[H]
\caption{Forward path: T-Net Layer}
\label{alg:tnet}
\begin{algorithmic}
    \REQUIRE input $\{\mathbf{X}, \mathbf{Y}\}$ 
    \ENSURE $\mathbf{\Tilde{z}}$
    \STATE $\mathbf{\Tilde{z}} \leftarrow \text{BNorm}_{0}(x)$  \# $b \times n \times m $
    \STATE $\mathbf{\Tilde{z}} \leftarrow \text{ReLU}  (\text{BNorm}_{1}(\text{Conv1D}_{1}(x)))$  \# $ b \times m \times d_{emb}$
    \STATE $\mathbf{\Tilde{z}} \leftarrow \text{ReLU}(\text{BNorm}_{2}(\text{Conv1D}_{2}(x)))$  \# $b \times d_{emb} \times 2\cdot d_{emb}$
    \STATE $\mathbf{\Tilde{z}} \leftarrow \text{ReLU}(\text{BNorm}_{3}(\text{Conv1D}_{3}(x)))$ \# $b \times 2\cdot d_{emb} \times 4\cdot d_{emb}$
    \STATE $\mathbf{\Tilde{z}} \leftarrow \text{GlobalMaxPooling}(x)$  \# $ b \times 1 \times 4\cdot d_{emb}$
    \STATE $\mathbf{\Tilde{z}} \leftarrow \text{ReLU}(\text{BNorm}_{4}(\text{FFN}_{1}(x)))$  \# $b \times 1 \times 2\cdot d_{emb}$
    \STATE $\mathbf{\Tilde{z}} \leftarrow \text{ReLU}(\text{BNorm}_{5}(\text{FFN}_{2}(x)))$ \# $b \times 1 \times d_{emb} $ 
    \RETURN $\mathbf{\Tilde{z}}$
\end{algorithmic}
\end{algorithm}
The components of the T-Net layer involved can be described as follows:
\begin{itemize}
    \item $\text{Conv1D}_l$: Signifies the application of different 1D-convolution layers with progressively increasing size ($d_{emb},2\cdot d_{emb}, 4\cdot d_{emb}$).
    \item $\text{BNorm}_l$: This layer conducts batch normalization to enforce the output's zero mean and unit variance distribution. The function comprises five distinct instances of such units due to learnable parameters that facilitate a shift in the output.
    \item ReLU: After calling the particular layer and normalizing the result, the Rectified Linear Unit (ReLU) is applied to introduce sparsity. 
    \item GlobalMaxPooling: This aggregation determines the maximum values across a dimension, which ensures a reduction and provides invariance.
    \item $\text{FFN}_l$: Before obtaining the result, the information is compressed from size $(1 \times 2\cdot d_{emb})$ to $(1 \times d_{emb})$ by passing through a two-layered feed-forward neural network (FFN).
\end{itemize}
\noindent\textbf{Decoder}: As previously stated, the decoder component takes the latent information and previously determined sequence ($\mathbf{s}$) as input to calculate a matrix ($\textbf{P}$). This matrix comprises the log probabilities corresponding to each viable candidate for the subsequent symbol at the $n$-th position in the sequence:
\begin{equation}
\label{eqn:prop-vec}
\textbf{P}=D(\mathbf{\Tilde{z}}, \mathbf{s};\theta_D) 
\end{equation}
with $\theta_D$ representing the parameters. In contrast to the conventional transformer decoder architecture as described by Vaswani et al. \cite{Vaswani2017AttentionNeed}, we utilize a modified version of the open and efficient foundation language model (LLaMA) proposed by Touvron et al. \cite{Touvron2023LLaMA:Models}. Including adaptations such as RMSNorm-based pre-normalization \cite{Nguyen2019TransformersSelf-Attention}, the SwiGLU activation function \cite{Shazeer2020GLUVI}, and the rotary embedding mechanism \cite{Su2021RoFormerET}, the LLaMA model demonstrates superior performance over models like GPT-3 on various benchmarks \cite{Touvron2023LLaMA:Models}. Notably, this is achieved by only using 4\% to 30\% of the parameter amount of GPT-3. With our adaptation, i.e., implementing a cross-attention layer to infer $\mathbf{\Tilde{z}}$, the decoder consists of the following elements:
\begin{itemize}
\item Tokenembedding: The function of the token embedding converts the $\mathbf{s}$ into a continuous floating point representation ($\mathbf{\hat{s}}$) with a dimension of the context window $\times d_{emb}$.
\item RMSNorm: Analogous to the encoder's batch normalization, this component modifies the output to a statistical mean of zero. The specific preference for using RMSNorm over classic layer normalization is due to its computational efficiency \cite{zhang2019root}. 
\item $\text{Attn}_l$: Within this function, the multi-head attention layer is applied with respect to the embedded input sequence ($\text{Attn}_1$) and in terms of the cross-relation ($\text{Attn}_2$) between the embedded sequence and the $\Tilde{z}$, determining the contextual relation between the representation of the symbols and the latent vector. 
\item FFN: This component signifies a two-layer feed-forward network with a layer size of $4\cdot d_{emb}$ and $d_{emb}$ respectively.  
\end{itemize}
Except for the token embedding function, the operations described herein comprise what is commonly referred to as a transformer block. This component consists of several operations, excluding the token embedding function, which enables information propagation through multiple blocks. Algorithm (\ref{alg2:tblock}) summarizes the steps involved in processing a single instance that takes in the embedded sequence ($\mathbf{\hat{s}}$) and the latent information. The method starts by invoking normalization (RMSnorm), followed by a self-attention function call (lines 1-2). The result from employing a normalization on the residual connection ($\text{res}_1$) is in addition to $\mathbf{\Tilde{z}}$ an input parameter for the cross-attention operation (lines 3-4). Repeating the procedure of passing the result ($\text{att}_2$) through a residual connection with the inference of the fully connected network ($\text{FFN}$) provides the transformed variable for the next block or serves as the output (lines 5-7). For a more detailed understanding of the individual elements, please refer to \cite{Touvron2023LLaMA:Models} or \cite{Vaswani2017AttentionNeed}.
\begin{algorithm}[H]
\label{alg2:tblock}
\caption{Forward path: transformer block}
\begin{algorithmic}[1]
\REQUIRE Inputs $\mathbf{\hat{s}},\mathbf{\Tilde{z}}$
\ENSURE $\mathbf{\hat{s}}$; $\mathbf{\Tilde{z}}$
\STATE $\text{x} \leftarrow \text{RMSNorm}(\mathbf{\hat{s}})$
\STATE $\text{att}_1 \leftarrow \text{Att}_1(\text{x},\text{x}) $
\STATE $\text{res}_1 \leftarrow \text{RMSNorm}(\text{x} + \text{att}_1)$
\STATE $\text{att}_2 \leftarrow \text{Att}_2(\text{res}_1,\mathbf{\Tilde{z}})$
\STATE $\text{res}_2 \leftarrow \text{RMSNorm}(\text{res}_1 + \text{att}_2)$
\STATE $\text{x} \leftarrow \text{FFN}( \text{res}_2 )$
\STATE $\mathbf{\hat{s}} \leftarrow \text{res}_2+ \text{x}$
\RETURN $\mathbf{\hat{s}}$; $\mathbf{\Tilde{z}}$
\end{algorithmic}
\label{alg:custom_transformer_layer_call}
\end{algorithm}
\subsection{Sampling} 
\noindent The sampling mechanism describes a procedure for retrieving a token given by a probability vector, which results from the decoder. Studies in natural language processing \cite{Vaswani2017AttentionNeed} show that it substantially affects the quality, variety, or perplexity of the sequence acquired. For creating the initial population with $n$ distinct individuals related to a problem or their building blocks, the quality is indicated by:
\begin{enumerate}
\item Algebraic consistency: The network training must ensure that computational operations are syntactic valid, which entails satisfying the minimum input parameters for symbols drawn from the nonterminal symbol set. This is crucial for preventing the generation of syntactically incomplete or incorrect token sequences, such as ``$x+y-*$".
\item Plausibility: As stated in \cite{Petersen2019DeepGradients}, the search space needs to be constrained to minimize the likelihood of obtaining nested combinations of trigonometric-, logarithmic- or exponential functions, for instance, $\sin(\cos(\log(x)))$. 
\end{enumerate}
To satisfy attribute one while maintaining the same training effort and network size, we leverage the capabilities provided by the Karva string (ref. equation (\ref{eq-seq_prop})). Converting this into a vector mask seamlessly integrates into the auto-regressive sequence generation process through a simple vector addition operation, thereby incurring minimal computational overhead. This mechanism also provides flexibility to hide different functions during the sampling process, satisfying criterion two. Taking into account the output of the transformer decoder, we obtain $\textbf{P}$, with a dimension of ($1 \times |\mathbf{C}|\times size_{w}$). Here, this number is specified by the batch size (which will be neglected in the subsequent consideration), the cardinality of the vocabulary $|\mathbf{C}|$ and the size of the context window. 

Given a sequence $\text{s}$ of length $t-1$ and $\textbf{P}$ of $D(\mathbf{\Tilde{z}_{t-1}}, \mathbf{s};\theta_D)$ the column at position $t$ provides the probability vector ($\textbf{p}_t$) for the next token. To ascertain the next symbol for the sequence, an initial manipulation of the probabilities is performed using the constructed mask, setting the log probabilities to $-\infty$:
\begin{equation*}
\hat{\textbf{p}}_{i} =
\textbf{p}_{i}+
\begin{bmatrix}
0 \\
0 \\
\vdots \\
0 
\end{bmatrix}
\text{for } t< H_{\text{len}} \\
\end{equation*}
\begin{equation*}
\hat{\textbf{p}}_{i}=  
\textbf{p}_{i}+
\begin{bmatrix}
-\infty \\
-\infty \\
\vdots \\
0 
\end{bmatrix}
\text{for } t \geq H_{\text{len}}\, .
\end{equation*}
Subsequently, the softmax function is applied to $\hat{\textbf{p}}_{i}$, transforming it into a probability distribution. The final step involves using the top (p) and top (k) sampling methods to select the next symbol in the sequence. The paradigm introduced via GEP inherently ensures the algebraic validity of the generated expression. Consequently, the decoder is relieved from the exhaustive task of explicitly learning to construct algebraically correct functions, thereby accelerating the learning effort.

\section{Experiment}
\label{results}
\noindent In this section, we compare the performance of the induced individuals created by the trained language model (IGEP) against the conventional method (GEP) on four data sets from the UCI database and an example from the area of computational fluid dynamics (CFD). The UCI data sets are chosen as test cases for the new framework for quantitative analysis, whereby the average performance across a multitude of test iterations is considered. The problem from the field of fluid dynamics, especially turbulence modeling, aims to ascertain whether the starting population also qualitatively corresponds to the problem solution. Here, table \ref{tab:cases} provides an overview of the cases in addition to their feature amount. The performance of the new framework is evaluated based on the following questions:
\begin{enumerate}
\item Is the selected dimension of the network and the knowledge of a small set of physical equations sufficient to influence the start point of the optimization positively?
\item What is the effect on exploration and exploitation in the evolutionary process?
\item Can we measure a difference between transferring individuals from the last generation or applying a language model?
\end{enumerate}
We start with a description of the particular configurations, encompassing details about the regression formulation, error metric, parameterization of the language model, training, and the software components employed. Following a summary of the results, we will elucidate factors contributing to the performance, identify some vulnerabilities, compare the computational overhead, and suggest a few ideas for further improvement.

\subsection{Experimental setup}
\noindent A crucial prerequisite to applying transfer learning is the existence of a representation pertaining to the solution or at least an approximation of $\mathcal{T}_s$. In the context of the proposed method, this representation encompasses two primary facets. Firstly, it consists of an ensemble of approximations denoted by $\Tilde{f}$, or an exact solution denoted by $f$. Secondly, it includes the input set characterized as $\{\mathbf{X}, f(\mathbf{X})\}$. Similar to the studies \cite{Dinh2015TransferProgramming} and \cite{Munoz2020TransferProgramming}, the ensemble of possible solutions is provided through different trials using the non-deterministic nature (obtaining a slightly different solution for each trial) of the evolutionary method. Instead of exclusively using individuals from the last generation, who are likely to be overfitted, we consider a proportion from the whole course of the exploration. Moreover, the analogy to the pre-trained approaches provides a decoupling of the direct projection of knowledge from $\mathcal{T}_s$ to $\mathcal{T}_t$. Referring to our test scenarios, the configurations are defined as follows:
\\\\
\textbf{Configuration for the UCI test cases}\\
We select eight cases from the overall UCI database, listed in table \ref{tab:cases} (1-8) with their given feature range. These can be described using the typical formulation for a regression problem:
\begin{equation}
\hat{f}(\mathbf{X}) = \mathbf{\hat{y}} \approx \mathbf{y}\,.
\end{equation}
The minimization is performed by successively evaluating precision of $\hat{y}$, using the mean absolute error (MAE):
\begin{equation}
\label{eqn:error_metric_uci}
   fitness\equiv\text{MAE}(\mathbf{y},\mathbf{\hat{y}}) = \frac{1}{|\mathbf{y}|}\sum_n^{|\mathbf{y}|}|\hat{y}_n-y_n| \, .
\end{equation}
Assuming that established physical relations offer a favorable outset for tackling subsequent problems, we employ the Feynman benchmark equation set \cite{matsubara2023rethinking} as $\mathcal{T}_s$. Consequently, a training data set for the language model is generated by executing our evolutionary method across 200 generations within five runs. This results in a diverse ensemble of representations to approximate the given problems, whereby the feature dimension is limited to eight. The sequences subsequently undergo additional post-processing to filter out individuals exhibiting nested expressions and possessing a fitness close to the explored minimum ($15\%$ larger than the minimum), resulting in approximately $5\cdot10^4$ samples. In the final stage, the sequences $\mathbf{S}$ are tokenized using the vocabulary extracted from the equations and subsequently converted into tensors of fixed length 64. Any unoccupied positions within these tensors are padded with zeros to ensure dimensional consistency. The model, notably the components $\theta_\mathcal{E}$ and $\theta_D$ to be employed, displays the following attributes:
\begin{itemize}
\item $\theta_D$: For the decoder, we adopt an embedding dimension of 32, eight attention heads within ($Attn_1$, $Attn_2$), eight transformer blocks, and a context window size of 64.
\item $\theta_\mathcal{E}$: The parametrization of the encoder is predicated on the specified embedding size, culminating in a configuration as delineated in Algorithm \ref{alg:tnet}.
\end{itemize}
To facilitate effective knowledge transfer, the features from forthcoming considerations are mapped into the established vocabulary predicated on their inherent units.\\\\
\begin{table}[!t]
\caption{Experimental data set}\label{tab:cases}
\centering
\begin{tabular}{|c|c|c|c|}
\hline
Nr.&Target data set  & Input features & Target value\\
\hline
1&Gas Turbine \cite{misc_gas_turbine_co_and_nox_emission_data_set_551} & 9 & Nitrogen oxides\\
&CO and NOx&&\\
\hline
2&Cycle Power Plant\cite{misc_combined_cycle_power_plant_294}& 4 & Energy output \\
&Emission data set&&\\
\hline
3&Compressive Strength\cite{misc_concrete_compressive_strength_165} & 8 & Compressive\\
&&&strength\\
\hline
4&Airfoil Self-Noise\cite{misc_airfoil_self-noise_291} & 5 & Sound pressure \\
\hline
5&Real Estate Valuation\cite{misc_real_estate_valuation_477} & 6 & Price \\
\hline
6&Energy Efficiency\cite{misc_energy_efficiency_242} & 8 & Heating load \\
\hline
7&Yacht Hydrodynamics\cite{misc_yacht_hydrodynamics_243} & 6 & Residuary resistance \\
\hline
8&Computer Hardware\cite{misc_computer_hardware_29} & 10 & CPU performance \\
\hline
9&Duct Flow \cite{Weatheritt2016} & $>$5 & Anisotropic \\
&Velocity related change&&Reynolds stress \\
\hline
10&Duct Flow \cite{Weatheritt2016} & $>$5 & Anisotropic \\
&Domain related change&&Reynolds stress \\
\hline
\end{tabular}
\end{table}
\begin{table}[!t]
\caption{Experimental Parameters\label{tab:paras}}
\centering
\begin{tabular}{|c|c|c|}
\hline
Property  & UCI test cases & CFD problem \\
\hline
Population size & 100 & 100\\
\hline
Sampled proportion & 0.1,0.25,0.5,0.75 & 0.6\\
\hline
Head length & 6 & 5\\
\hline
Number of genes &3& 4 $\times$ 3 \\
\hline
Reproduction rate & 0.9 & 0.1\\
\hline
Mutation probability &  0.05 & 0.05\\
\hline
One-point crossover  & 0.3 & 0.5\\
probability &&\\
\hline
Two-point crossover  & 0.2 & 0.4\\
probability&&\\
\hline
Function set & $\{+,-,*,/, sqr,sqrt\}$ & $\{+,-,*\}$\\ 
\hline
Max. generations& 100 & 30\\
\hline
Number tokens ($c$) & $\{0.1,0.5,-0.1,-0.5\}$ &$\{0.1,0.5,-0.1,-0.5\}$ \\
\hline
\end{tabular}
\end{table}
\begin{table*}[t]

\caption{Average $\min(fitness)$ after 100 generations on 25 different runs, here listed in comparison to the conventional \textbf{GEP}, the GEP with the augmented start population (\textbf{IGEP}), the \textbf{gplearn} package, the \textbf{DSR} approach and the GP-GOMEA.}
\centering
\begin{tabular}{c|ccccc|ccc}
\hline
\label{table-Comparison}
Case-Nr. & $\mathbf{GEP}$ & $\mathbf{IGEP_{0.1}}$ & $\mathbf{IGEP_{0.25}}$ & $\mathbf{IGEP_{0.5}}$ & $\mathbf{IGEP_{0.75}}$ & $\mathbf{gplearn}$ & $\mathbf{DSR}$ &$\mathbf{GP-GOMEA}$ \\
\hline
1 & $6.33$ & $\mathbf{5.07}$ & $5.12$ & $5.18$ & $5.05$ & $8.78$ & $11.15$  & $7.09$\\
\hline
2 & $5.61$ & $5.53$ & $4.56$ & $4.85$ & $6.30$ & $7.01$ & $79.95$ & $\mathbf{3.71}$ \\
\hline
3 & $9.18$ & $7.15$ & $7.18$ & $7.19$ & $\mathbf{7.10}$ & $8.76$ & $16.60$ & $9.06$\\
\hline
4 & $10.99$ & $8.50$ & $8.09$ & $7.35$ & $9.93$ & $27.60$ & $31.62$ & $\mathbf{4.01}$\\
\hline
5 & $0.22$ & $0.19$ & $\mathbf{0.09}$ & $0.19$ & $0.24$ & $10.39$ & $10.63$ & $6.11$ \\
\hline
6 & $1.17$ & $1.18$ & $\mathbf{1.16}$ & $1.18$ & $1.18$ & $3.27$ & $5.22$ & $2.74$\\
\hline
7 & $2.27$ & $2.21$ & $\mathbf{2.14}$ & $2.59$ & $2.52$ & $2.65$ & $7.83$ & $3.56$\\
\hline
8 & $22.05$ & $22.14$ & $\mathbf{21.91}$ & $22.34$ & $22.01$ & $22.07$ & $26.20$ & $22.18$ \\
\hline
\end{tabular}
\end{table*}

\noindent\textbf{Configuration for the CFD scenario}\\
The configuration for this problem category differs from the symbolic regression applications investigated above in terms of the dimensionality of the algebraic structures involved in the regression. Here, a tensor regression is sought, while the conventional test cases deal exclusively with scalars. Hence, referring to proposed methodology of Weatheritt and Sandberg \cite{Weatheritt2016}, the problem is formulated as follows: 
\begin{equation}
\label{eqn:ai}
a_{ij} = \sum_k(\lambda_k(\mathbf{X}_n)\cdot T_{ij}^k)\,,
\end{equation}
where $\lambda_k$ denotes a spatially varying coefficient, and $T_{ij}^k$ signifies a basis tensor. Flow predictions are obtained by incorporating the correction into a system of partial differential equations and executing a numerical simulation to solve the discretized equations over a physical domain. Here, we focus on obtaining the velocity field ($\mathbf{U}$):
\begin{equation}
\mathbf{U}=f(\phi_p, a_{ij},...)\,,
\end{equation}
with $\phi_p,$ denoting a set of further physical quantities. The error of the model prediction, or low-resolution result ($l$), is then computed by comparing it to a snapshot from a high-fidelity simulation (high resolution - $h$) or measurements from experiments over the spatial domain, here ($y,z$), by a weighting with respect to the high-fidelity quantities:
\begin{equation}
  fitness\equiv\ \int (||\mathbf{u}_{2:3}||_{h} - ||\mathbf{u}_{2:3}||_{l} ) \cdot w_h  \,\mathrm{d} y \, \mathrm{d} z \, . 
\end{equation}
The notation $|{u}_{2:3}|$ signifies the vector components along the $y$ and $z$ directions. A more detailed description of this case is given in the Appendix. Given the composition of the approximation and the sensitivity of this correction to variables such as initial velocity or domain size, a more targeted projection between $\mathcal{T}_s$ and $\mathcal{T}_t$, similar to the approach in \cite{Dinh2015TransferProgramming}, is adopted. Consequently, the training data set is derived from a canonical case characterized by a $z/y$ ratio of 1 and a friction Reynolds number of 180 ($Re{\tau}$, a metric representing the ratio between inertial and viscous forces within the fluid flow). Through the execution of 10 CFD simulations with different corrections $a_{ij}$ per generation, over 200 generations, a minimum fitness of approximately $0.14$ for $\mathcal{T}_s$ is obtained. In this setup, each simulation takes about 90 seconds, whereby in general, this duration depends on factors like the problem complexity, the domain size, and the resolution. The post-processing of the sequence and the tokenization is employed similarly to the UCI test configuration with an extension of the window size to 256. The dimension of the task necessitates adaptations to $\theta_\mathcal{E}$ and $\theta_D$:
\begin{itemize}
\item $\theta_D$: The proposed changes for the decoder mainly affect the embedding size, which is extended to 128 and the number of transformer blocks is increased from 8 to 12. 
\item $\theta_\mathcal{E}$: The encoder inherits the dimensions from the defined embedding size. Additionally, the input for the encoder (Eqn. \ref{eqn-encoder}) changes from $(\mathbf{X},\mathbf{Y})$ to $(\mathbf{X})$, reflecting the absence of the target values $(\mathbf{Y})$.
\end{itemize}
The configurations undergo a training of 50 (first configuration) and 500 (second configuration) generations, employing the AdamOptimizer \cite{kingma2017adam} alongside a weight-decaying learning rate scheduler. This procedure is conducted over mini-batches, each comprising 128 data records for the UCI scenario and 256 for the flow scenario. The parameters utilized for the evolutionary method are delineated in Table \ref{tab:paras}. The setup is executed using the GEP-python implementation 'geppy' \cite{geppy_2020} for cases (1-8), the EVE-implementation \cite{Weatheritt2017TheAlgorithm} for case (9-10), and a language model extension encapsulated in TensorFlow (v.2.11.0). The computational resources used include an NVIDIA A100 for training the neural network and a server with 32 cores (2.0GHz) for executing the exploration using the new framework.

The described experimental scenarios maintain a pool of 100 candidate solutions, of which, for the first configuration, a proportion ranging from 0.1 to 0.75 is composed by the language model and introduced into the initial population. The selection of additional hyperparameters for the UCI data is guided by recent studies \cite{Kasten2022AnAlgorithms} and \cite{Zhang2021a}. 
Given that the optimization requirements for the flow scenario are different from the UCI cases, particularly in terms of complexity, we slightly modify the parameters and only consider a sampled proportion of 0.6. The change in the recombination rate is mainly dictated by the necessity for a numerical simulation to compute the error.
\subsection{Result and Discussion}
\noindent
The initial segment of our comparative analysis focuses on assessing the impact of knowledge transfer in approximating expressions for the mentioned UCI cases. We specifically evaluate the conventional GEP and the GEP creating an inductive start population (IGEP) across 25 different runs, employing Eqn. \ref{eqn:error_metric_uci}. These are compared to three established methods from the literature to assess the relative performance, including the gplearn \cite{stephens2015gplearn}, the DSR \cite{Petersen2019DeepGradients}, and GP-GOMEA proposed by Virgolin \cite{virgolin2021improving}. The hyperparameters have been chosen similar to table \ref{tab:paras}, apart from the DSR, where a sample size of $10^6$ was selected.

Table \ref{table-Comparison} summarizes the average minimum fitness achieved in each experiment. Notably, every case that employed an augmented proportion of 25\% shows an improvement compared to the standard GEP. However, the IGEP configurations with 50\% and 75\% augmentation underperform in specific cases, e.g., $2.59$ in case seven and $0.24$ in case (5), compared to the standard method. A proportion of 10\% leads in the worst case to a result close to the standard procedure. This suggests that the augmented expressions may not always align well with the data set characteristics, revealing potential areas for further refinement of the language model's training data set.

In comparison to the DSR, the GEP generally performs better, even without the augmented individuals, indicating that the gradient policy might need far more samples to reach a lower error. The gplearn package shows a better result than standard GEP in case (3) ($8.76$ against $9.18$), whereby the augmentation of the start population in IGEP overcomes this gap. In cases of finding an expression to predict the energy output (2) or the airfoil-self-noise (4), the  GP-GOMEA clearly outperforms the GEP even when using transferred knowledge. This performance can be related to the capabilities of the included techniques, implying that additional mechanisms during the exploration, like semantic backpropagation, are able to compensate for the absence of an augmented start population.

With regard to the creation process, we focus on the first four cases from table \ref{tab:cases} and delve into the impact of varying sampled proportions, where a notation of '0' signifies the application of the standard method. A further indication of a positive influence of using sampled individuals rather than a random initial population can be estimated by the fitness within the start population.
\begin{figure*}[!t]
\centering
\includegraphics[width=6.5in, height=2in]{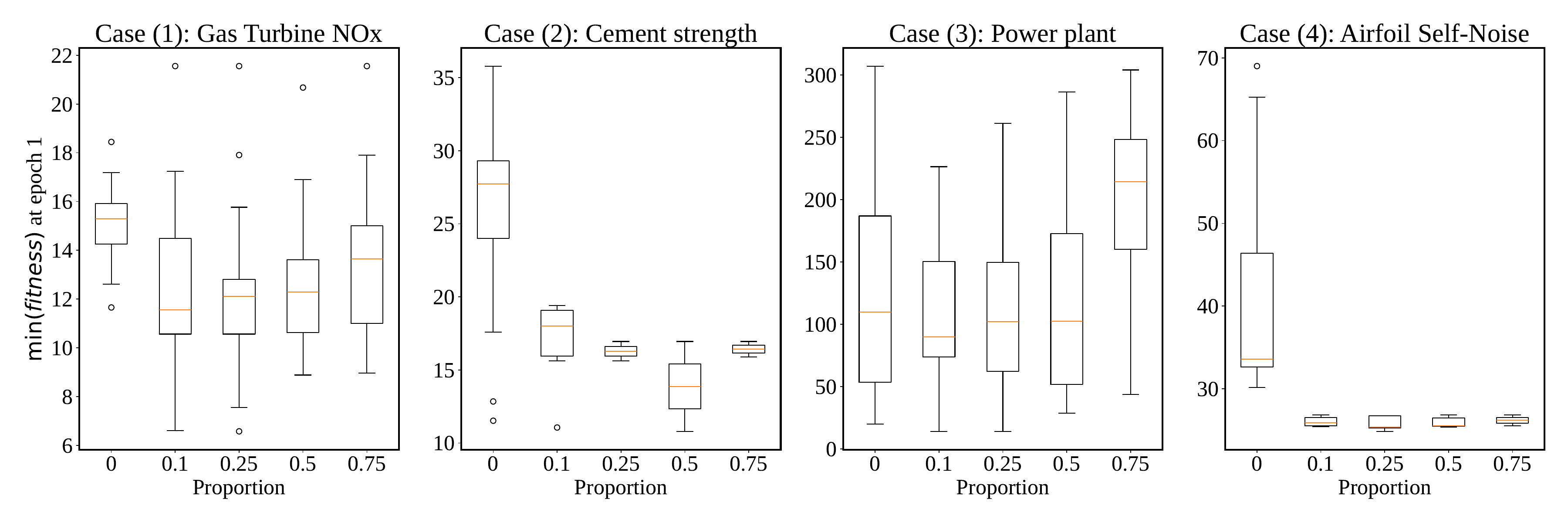}
\caption{Boxplot illustrating the distribution of the minimum error over the 25 performed runs. Here, comparing the conventional application (GEP - sampled 0) after the initial generation with the proposed method, considering different proportions of augmented individuals $(0.10,0.25,0.50,0.75)$ on four different cases from the UCI database. A lower fitness signifies a more suitable approximation.}
\label{fig_1:start_error}
\end{figure*}

Fig. \ref{fig_1:start_error} offers the fitness distribution of the best individuals for cases (1-4) obtained after the initial generation, separated by a different proportion of sampled individuals. For cases (2) and (4), a significantly smaller error can be observed on average for all proportions, quantifiable at 32\% (2) and 30\% (4). Reasons include that $\mathcal{T}_t$ shows similarity to $\mathcal{T}_s$, the chosen attributes and the number of features are sufficient for the problem description, and the mapping of the units of the individual columns is unique. Additionally, the application of constructed solutions results in a higher dispersion. However, task (3) does not exhibit any notable influence for any proportion ranging from $0.10$ to $0.50$ or even shows a negative impact. Here, a substitution of 75\%  results in a shift of the median fitness from approximately $126$ to roughly $200$. This observation, especially in the context of cases (2) and (4), suggests that extrapolating to higher dimensions beyond those covered in $\mathcal{T}_s$ significantly amplifies the likelihood that the constructed individuals will exert no positive influence on the solution. Furthermore, it is evident that inference is incapable of compensating for the insufficient quality and number of predictors.
However, with the exception of this case, no unique characteristic emerges for the different proportions investigated, which, on average, show similar performance.

\begin{figure*}[!t]
\includegraphics[width=6.5in, height=2in]{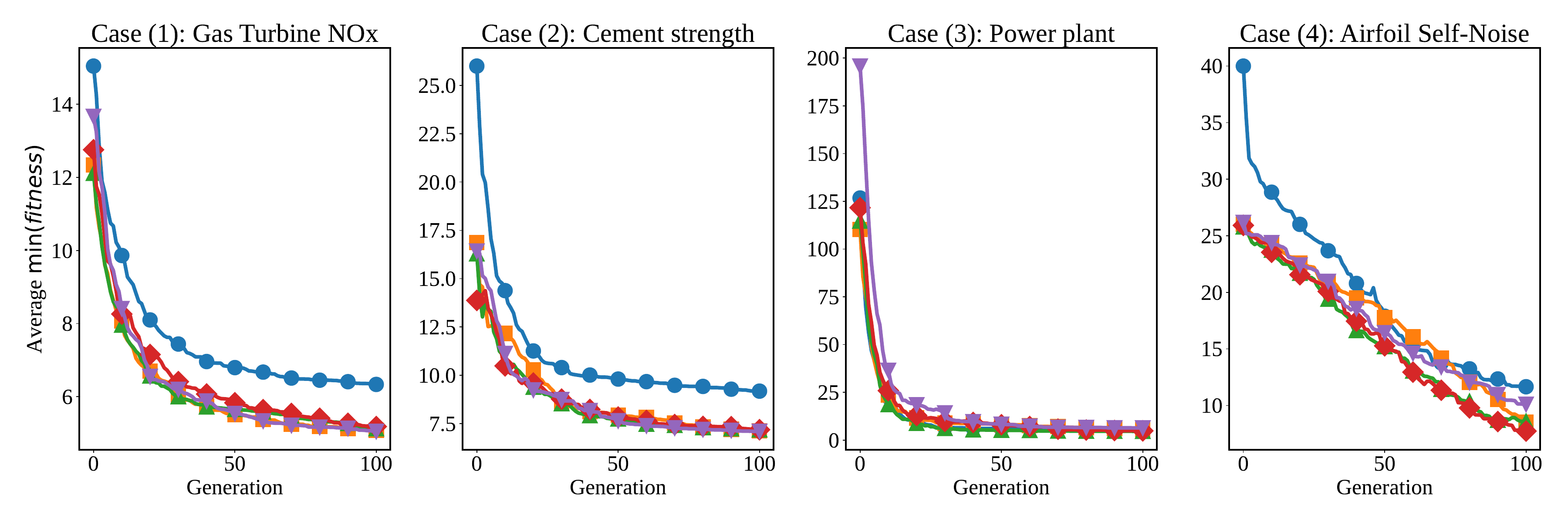}
\centering
\includegraphics[width=3in, height=0.5in]{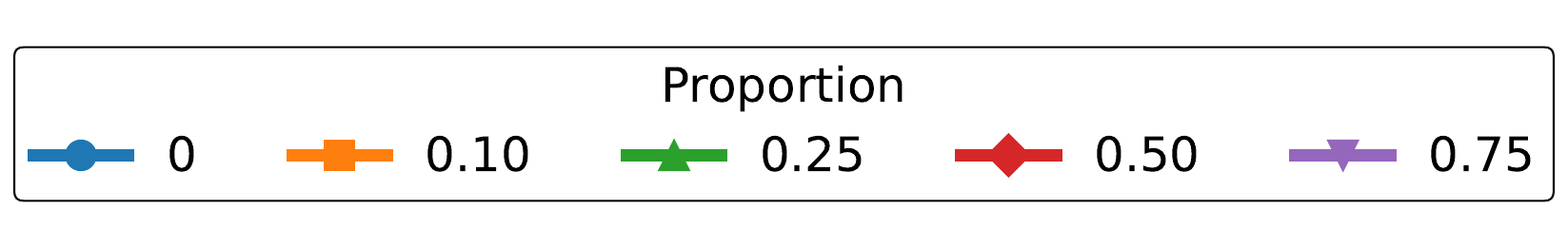}
\caption{Diagram presenting the average MAE of the fittest individuals for 25 different trials over 100 generations, whereby each plot shows the result for a particular case. The different markers indicate the proportion according to the sampled individuals, ranging from zero (standard GEP) over $0.1,0.25,0.5$ to $0.75$.}
\label{fig_2:eval_epoch}
\end{figure*}

In conjunction with the start performance, Fig. \ref{fig_2:eval_epoch} offers an aggregated insight into the exploration behavior of the individual cases employing the operators of the evolutionary method. The average minimum fitness over the different test runs continues the initial trend, slightly reduced within subsequent iterations. In a broad comparison between the standard method and the induced method, we observe that the former starts with weaker fitness values but is able to undertake a very significant correction within the first 20-40 generations. The latter, on the contrary, starts with a smaller error but cannot sustain the initial fitness difference. 

This issue arises from the disruptive behavior exhibited by the genetic operators, which may inadvertently transform combinations that are potentially advantageous into configurations with diminished effectiveness \cite{Haslam2016}. In the cases of NOx and cement strength prediction, the relaxation towards the minimum exhibits a decrease after approximately 20 and 35 generations, respectively, for the standard method. Contrarily, in instances with adjusted start populations, this behavior is marginally attenuated, leading to a lower fitness for the conducted test cases: around 15\% for case 1 and 20\% for case 2. The average minimum fitness value achieved for a proportion of zero after 100 iterations is already achieved after 28 and 22 iterations during the tests with non-zero proportions. Consequently, referring to table \ref{tab:paras} and including the cumulative evaluations needed over the iterations, this corresponds to about 6,700 and 6,300 additional fitness evaluations required to achieve the same value. 

The exploration to predict the energy output of the power plant reflects a similar pattern to the dispersion of the error within the first generation. In this case, the insertion of individuals, with the exception of the negative transfer, results in a slight discrepancy, ranging from a fitness of 4.6 (proportion 0.25) over 5.6 (proportion 0) to 5.5 (proportion 0.1). For case (4), there is an additional positive trend due to the influence of the knowledge transferred. In addition to the expected lower fitness at the beginning for all transferred cases, we obtain lower fitness values in comparison (22\% for 0.75 up to 36\% for 0.5) to the baseline method via an almost linear progression. The expected fitness value of about 10.9 from the standard method is obtained on average from the explorations of the populations '0.25' and '0.50' in generation 72, thus with 28 fewer generations. 

The results presented for cases (1-4) indicate that the selected architecture, when applied in the prepared training state, is adept at generating building blocks from the given input. This leads to a beneficial starting point in the evolutionary process, indicated by lower initial fitness values. Additionally, on average, it achieves an optimal solution more quickly, i.e., with fewer generations. 

The identified weaknesses in this scenario primarily arise from an increasing dimensionality, a known challenge in pre-trained approaches \cite{Kamienny2022End-to-endTransformers}, coupled with the intrusion of disruptive operators, which essentially neutralize the benefits of augmented individuals, similar to observations in \cite{Haslam2016}. Moreover, there exists a likelihood of substantial dispersion and adverse influence if the problem is inadequately defined in terms of given predictors. Possible solutions include a mechanism that measures the uncertainty for the expression prediction and avoids inference when the case representation differs too much from the trained state of the language model or a technique to refine the predicted sequence based on the feedback of the error from the first generation.

The qualitative analysis, utilizing direct projection from $\mathcal{T}_s$ to $\mathcal{T}_t$, aims to ascertain the scope to which the augmented individuals are distinct with respect to the optimum. Moreover, it facilitates a direct comparison between an induction based on the best subtrees (CGEP), as described in \cite{Dinh2015TransferProgramming}, compared to the capabilities of the language model-based method (IGEP).

\begin{figure}
\centering
\includegraphics[width=2.75in, height=4.125in]{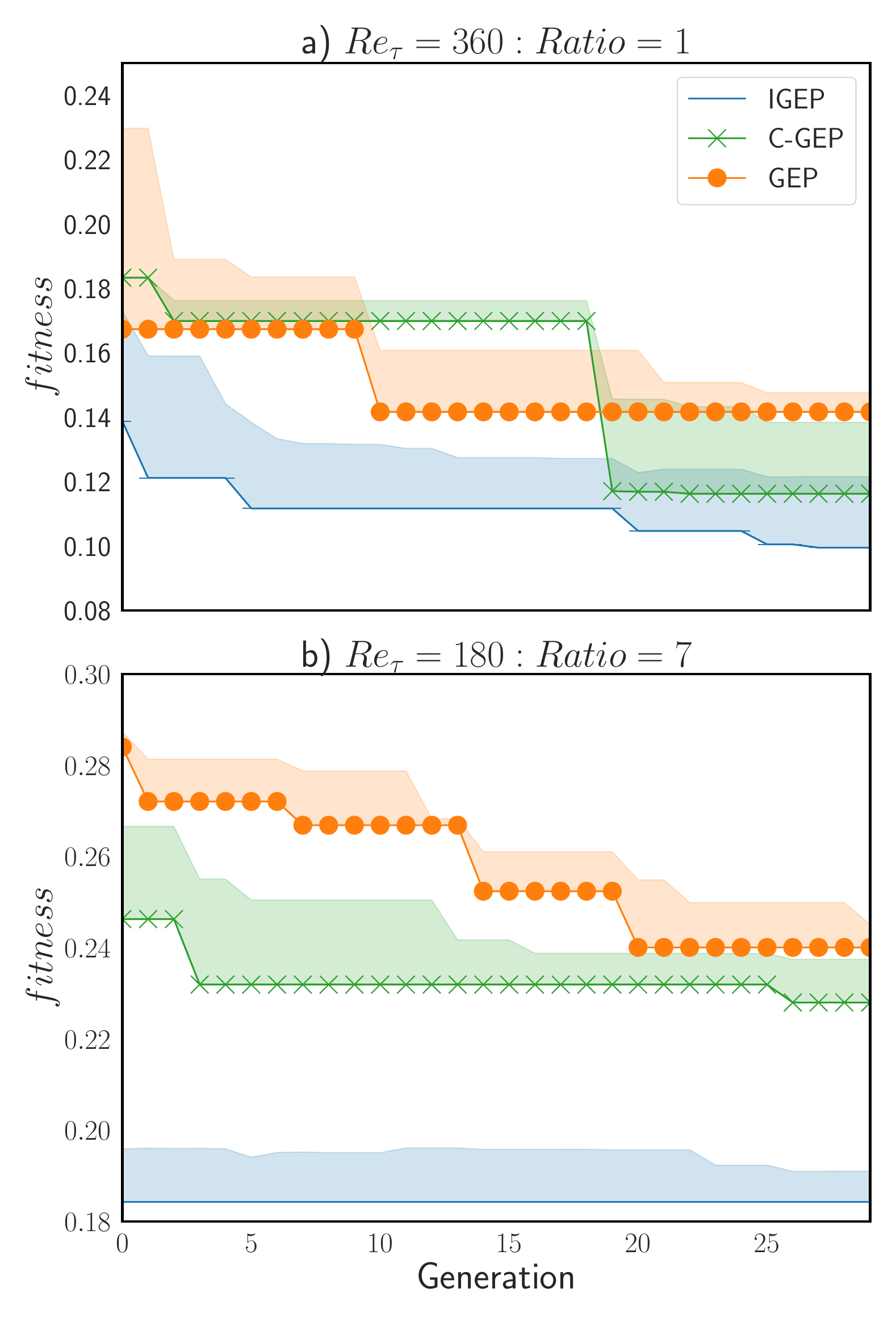}
\caption{Diagram illustrating the comparative exploration of the best individual in addition to their standard deviation across 10 different trials (as measured by fitness) over 30 generations for the two distinct target tasks within the flow scenario. The set of methods tested comprises the conventional procedure (GEP: 'o'), a technique that replicates the best subtrees from the source task (CGEP, 'x'), and the novel approach discussed here (IGEP: '-'). The target tasks are delineated as follows: a) relates to a change in velocity, and b) pertains to a change in the spatial domain.}
\label{fig_3:epochs_duct}
\end{figure}

Fig. \ref{fig_3:epochs_duct} shows a comparison of the two target tasks, where `$Re_{\tau}=360: Ratio=1$' (Fig. \ref{fig_3:epochs_duct}-a) denotes an increase in velocity, or inertial force relative to viscous force, while `$Re_{\tau}=180: Ratio=7$' (Fig. \ref{fig_3:epochs_duct}-b) signifies a change of the spatial domain, i.e. a different aspect ratio. The fitness assessment for the best candidate, as depicted by the standard deviation of the best results from additional trials, spans 30 generations and encompasses the three different methods. These include the conventional GEP, a replication of the best individuals from $\mathcal{T}_s$ (CGEP), and the introduced approach (IGEP). This observation further highlights the challenge associated with optimizing expressions for nonlinear environments. 

Despite achieving a reasonable fitness of 0.14 for the optimization task of $\mathcal{T}_s$ with $Re_{\tau}=180: Ratio=1$, simply replicating it (CGEP) for scenario (a) results in a fitness value that is worse ($10\%$) than what is obtained inferring 100 $\times$ 10 randomly initialized individuals (GEP). Individuals induced by the IGEP, as indicated by the deviation, only attach the level of random initializations in the most unfavorable scenario. Conversely, in the best-case scenario, the method generates a solution that standard GEP cannot achieve even after 30 generations.
In the case of a spatial change of the problem (\ref{fig_3:epochs_duct}-b), the IGEP outperforms the two comparative methods right from the beginning, with the improvement being about 25\% against the CGEP and 35\% against the GEP. Additionally, the information explored within $\mathcal{T}_s$ exhibits lower sensitivity to a spatial change, which subsequently results in a more gradual optimization curve for CGEP and stagnation for IGEP. This behavior may be attributable to the high proportion of induced candidates, significantly reducing variability in distinct expressions. Possible drawbacks include a negative transfer, which is somewhat mitigated in this context due to the similarity between the source and target tasks, or early convergence to a local minimum.

\subsection{Computational Overhead}
\noindent
By applying the methods GEP, IGEP, DSR, gplearn, and GP-GOMEA, we can measure an individual runtime ($t_r$) per method for each case, defined as the duration from the start of the optimization to the fulfillment of the termination criterion. Next to the underlying computing hardware utilized for the execution, properties highly affecting this quantity are the efficiency of the chosen programming language, the time required to evaluate the fitness ($t_f$), the number of evaluations per generation, or the application of further inner methods like semantic backpropagation and gene pool mixing within GP-GOMEA, the reward-based optimization for the recurrent neural network within the DSR, or the generation of the initial population via querying a language model according to the IGEP. In addition to $t_r$, the proposed method also requires a time for the preparation of the data set and the training of the language model ($t_p$), with $t_{p(\text{IGEP})} \approx 3.6\cdot10^3s$.

Conducting the UCI-cases (1-8) table (\ref{tab:cases}), with $t_f\ll 1s$, the order according to the average runtime time was observed as follows: 
\begin{align*}
t_{r(\text{gplearn})}<t_{r(\text{GP-GOMEA})}<t_{r(\text{GEP})}<t_{r(\text{IGEP})}<t_{r(\text{DSR})}\,.
\end{align*}
Here, the conventional GP tends to be the fastest, which results from the fact that there is only a straightforward evaluation per expression tree. The GEP is slower based on the decoding overhead, while the IGEP loses further time for querying the language model. GP-GOMEA, despite being written in a more efficient language, increases the runtime through the amount of fitness evaluation. Since the aforementioned inner methods collect that information from every part of the tree, the number can be quantified by $2\times (\text{tree high}-2) \times \text{population size} \times \text{generations}$, resulting in $2.79\cdot10^5$ (compared to $9\cdot10^3$ for GP), when choosing four as a tree high and the parameters from table (\ref{tab:paras}). The DSR results in the largest runtime since it performs next to the training a step of adjusting the coefficients of each sampled expression using a second-order optimization method (Broyden–Fletcher–Goldfarb–Shanno algorithm). Since $(\text{population size} \times \text{generations} \times t_f)\ll t_{p(\text{IGEP})}$ (ref. to table \ref{tab:paras}) for all the methods, an increase in these parameters does not significantly affect $t_r$, yet it is likely to achieve a similar fitness value comparable to that of the IGEP.

In comparison, when performing the cases (9 \& 10) table \ref{tab:cases}, where $t_f \approx 140s$, an adjustment of the hyperparameters produces a remarkable impact on $t_r$. This suggests that similar fitness levels can counterbalance the initial training time of  $t_{p(\text{IGEP})}$, especially since $\mathcal{T}_s$ results from former optimizations. For the scenario presented, it can be summarized that the transfer learning approach, applied to the GEP-based symbolic regression, achieves comparable fitness values to the other strategies at least 30 generations earlier. This translates to a reproduction rate of 0.1 and a duration of approx. 140 seconds per case. Ultimately, this results in a time-saving of $4.2 \cdot 10^5$ seconds (approximately 11 hours).

\section{Conclusion \& Future work}
\label{conclude}
\noindent In this article, we proposed a strategy that facilitates efficient transfer learning in combination with Gene Expression Programming applied to symbolic regression. 
Specifically, a transformer architecture (with a small parameter size) was chosen, which interacts with the GEP, employing the genotype definition to guide the sampling process. This architecture can be adapted from a set of source tasks using a set of approximations. The acquired knowledge can then be projected directly or utilized as an inductive bias to generate a starting population for solving a target task, with or without direct relation.

The impact on the minimum fitness of the start population and the exploration of the evolutionary method was investigated through a series of experiments, utilizing data from the UCI database on the one hand and approximating a scenario pertinent to fluid dynamics on the other.
While the method might not address deficiencies or scale to much higher dimensions, the experiments conducted demonstrate that even minimally trained model architecture significantly enhances the initial population of a range of different tasks compared to a randomized approach or a subtree inference from a final generation. 

Future work on this framework includes architectural refinements, modified training methodologies, and more capable representations of the source task. Furthermore, the framework is not limited to symbolic regression. It could also be extended to create a start population for combinatorial problems or boolean tasks.
\section{Acknowledgements}
This research was supported by The University of Melbourne’s Research Computing Services and the Petascale Campus Initiative. The work was supported by a Melbourne Research Scholarship provided by the University of Melbourne.
\renewcommand*{\bibfont}{\footnotesize}
\printbibliography

\clearpage
{
\appendix[Flow scenario - case description]
\noindent The subsequent chapter provides supplementary reading for the considered flow scenario 'Duct Flow'. We start by outlining the governing equations applied to perform the simulation and highlight the challenge that arises from them. The second section highlights how this affects the particular scenario and what the correction tries to solve.  
\subsection{Governing equations}
The computational method used in the provided example is the Reynolds averaged Navier-Stokes (RANS). This approach facilitates the simulation of the flow by resolving the mean flow quantities. Fundamentally, RANS is based on decomposing the flow quantities into a mean- and fluctuating component and further averaging the governing equations, describing the physical phenomena. For modeling an incompressible flow, this results in the conservation of mass and the conservation of momentum into the following equations:
\begin{equation}
\label{stokes-incompressible-conservation-average}
    \frac{\partial \bar{u}_i}{\partial x_i} = 0 
\end{equation}
\begin{equation}
\label{stokes-incompressible-momentum-average}
\frac{\partial \bar{u}_i \bar{u}_j}{\partial x_i} = -\frac{\partial\bar{p}}{\partial x_i} + \frac{\partial }{\partial x_j} \left( \nu\frac{\partial \bar{u}_i}{\partial x_i} - \overline{u_i'u_j'} \right)
\end{equation}
where $\bar{u}$ signifies the mean velocity, $\nu$ the viscosity, $\bar{p}$ the mean pressure and $x$ belongs to a spatial coordinate. Due to the correlation of higher statistical moments of the velocity fluctuations, the product $\overline{u_i'u_j'}$ does not result in zero, leading to an unclosed term (in the literature referred to as the turbulence closure problem). A common modeling approach is given by the linear Boussinesq hypothesis, referred to as linear eddy viscosity model (LEVM): 
\begin{figure}[!t]
    \centering
    \includegraphics[scale=.3]{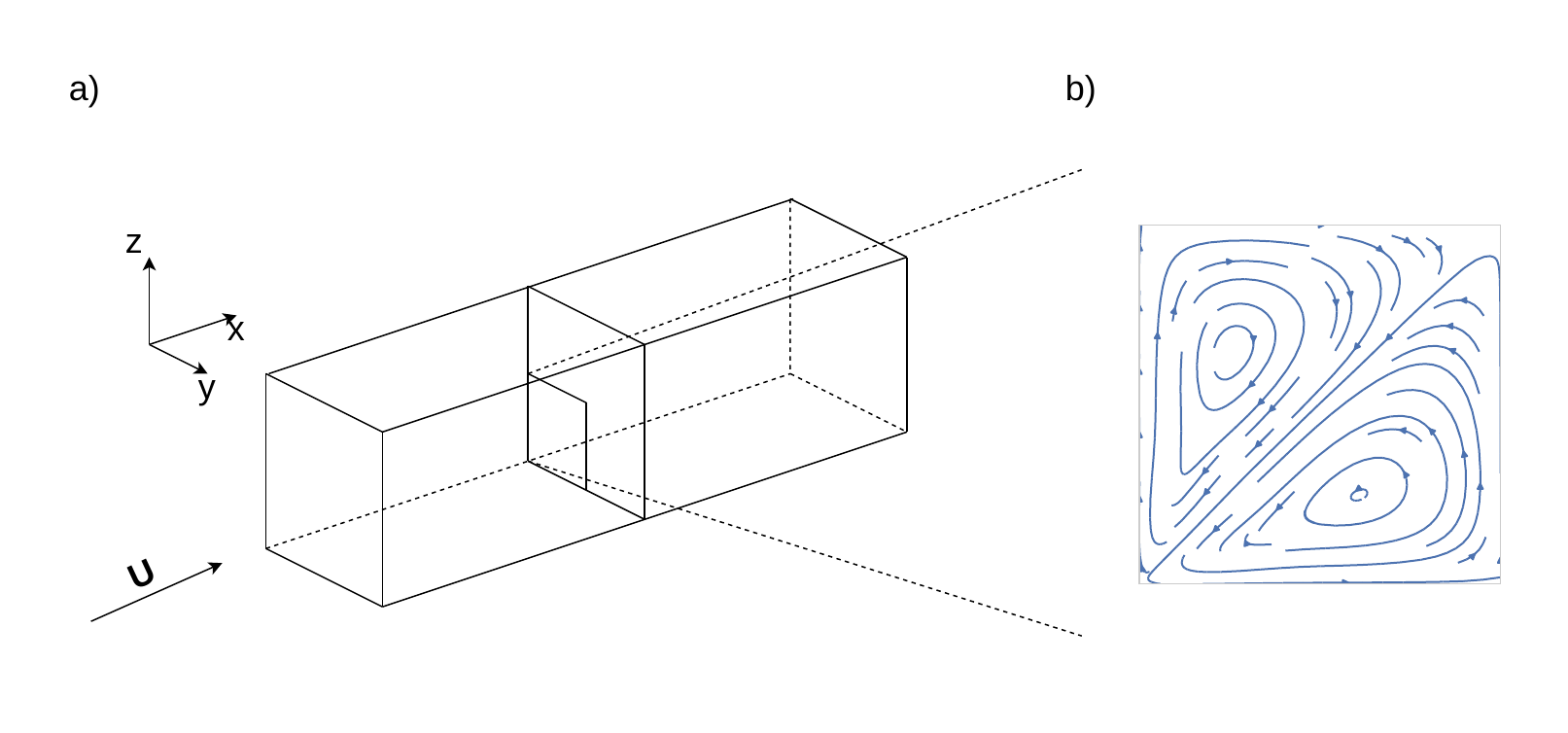}
     \caption{Duct case flow scenario with ratio one (a) and the secondary flow structure within the corner region (b) from a high-fidelity data set at $Re_{\tau}=180$.}
    \label{fig:duct_case}
\end{figure}
\begin{equation}
    \label{gl-levm}
    \tau_{ij} = \frac{2}{3}k\delta_{ij} - 2\nu_t \bar{S}_{ij} \, .
\end{equation}
Here $\tau_{ij}$ represents the Reynolds stress, $k$ is the turbulent kinetic energy, $\nu_t$ is the turbulent viscosity, and $\hat{S}_{ij}$ marks the mean strain rate tensor. The hypothesis given in equation (\ref{gl-levm}) defines a relation between the mentioned quantities and the required term but shifts the problem of inference to the effective viscosity ($\nu_t$), for which the literature shows a variety of possibilities of modeling for different phenomena and geometries (e.g. \cite{Menter1994}). Despite that assumption's capabilities, it lacks accuracy or capturing entire phenomena like a secondary flow. To obtain a more accurate estimate and to mitigate some of the limitations, a nonlinear isotropic term is added to $\tau_{ij}$ :
\begin{align}
    \label{gl-nlevm}
    \tau_{ij} = \frac{2}{3}k\delta_{ij} - 2\nu_t \bar{S}_{ij} + 2k a_{ij}^*\,.
\end{align}

\subsection{Flow Scenario}
\noindent The scenario describes an internal flow through a square duct as found in many engineering applications, such as heat exchangers, nuclear reactors, or ventilation systems \cite{Pirozzoli2017TurbulenceFlow}. A secondary flow phenomenon occurs in the corners (Fig. \ref{fig:duct_case}-b), which cannot be obtained by applying the linear eddy viscosity hypothesis. To approximate this phenomenon more precisely within a low-fidelity simulation, we use the GEP to explore $a_{ij}$. For creating the data of the source task, we start with the canonical case of a square duct, high ($y$) = wide ($z$), with $Re_{\tau}=180$. Fig. \ref{fig:duct_case}-a illustrates a small part of this object, where the arrow marked with $U$ denotes the direction of the mean flow. To save resources for performing $n$ simulations, each per expression that needs to be evaluated, we define the calculation domain as a quarter of the cross-section located in the corner (bottom-left) with no-slip boundary condition for boundaries next to the wall. The upper and right boundaries of that corner \ref{fig:duct_case} -a) are defined as symmetric. As a numerical solver, we employ the 'simple' implementation provided by the software OpenFoam (v.7)
} 
\end{document}